\documentclass[runningheads]{llncs}

\usepackage[T1]{fontenc}
\usepackage{graphicx}
\usepackage{amssymb,amsmath}
\usepackage{siunitx}
\usepackage{booktabs}
\usepackage{subcaption}
\usepackage{caption}
\usepackage[font=small]{caption}
\usepackage{tikz}
\usepackage{pgfplots}\pgfplotsset{compat=1.18}
\usepackage{placeins}
\usepackage[hidelinks]{hyperref}

\begin{document}


\title{Low-Magnification SEM May Suffice: Interpretable Deep Learning
for Multi-Scale Fracture-Cause Classification in
Zirconia-Toughened Alumina}

\titlerunning{Interpretable DL for Fracture-Cause Classification in ZTA}

\author{Julian Schmid\inst{1,2} \and
        Pawel Astankow\inst{1} \and
        Tom Vater\inst{1} \and
        Julius Beck\inst{1} \and
        Robert Cichon\inst{1} \and
        Danny Krautz\inst{1}}

\authorrunning{J. Schmid et al.}

\institute{CeramTec GmbH, CeramTec-Platz 1--9, 73207 Plochingen, Germany\\
\email{ju.schmid@ceramtec.de} \and
School of Life Sciences, University of Applied Sciences and Arts
Northwestern Switzerland, Hofackerstrasse 30, 4132 Muttenz, Switzerland}

\maketitle

\begin{abstract}
Reliable identification of fracture origins in alumina matrix composite hip and knee implants is critical for quality assurance and patient safety, yet current fractographic workflows are time-consuming, partly subjective, and reliant on high-magnification scanning electron microscopy (SEM). We present an interpretable vision-transformer (ViT) workflow for automated classification of fracture causes in an alumina matrix composite (BIOLOX\textsuperscript{\textregistered{}}\textit{delta}, CeramTec GmbH) widely used in total joint replacements. A dataset of 8{,}493 SEM images (50×--10{,}000×) was curated from five years of in-production burst and proof tests and annotated into three defect categories defined along the manufacturing chain: green body, hard machining, and material defects. Under severe class imbalance, the fine-tuned ViT reached an accuracy of 0.907 and a macro-F$_1$ of 0.888 in stratified five-fold cross-validation, with a two-stage perceptual-hash/SSIM leakage audit confirming negligible specimen overlap. Notably, performance at low magnification (50×) was comparable to that at high magnification (1k--10k×), indicating that macro-scale features---mirror geometry and hackle line fields---already encode sufficient diagnostic signal. Grad-CAM attributions consistently localised on canonical fractographic cues (mirrors, hackles, pores, machining marks), aligning with established fractographic criteria. Together, these results position interpretable ViTs as a complementary tool for ceramic-implant quality assurance, enabling low-magnification pre-screening and reducing reliance on time-intensive high-magnification inspection.

\end{abstract}

\section{Introduction}\label{sec:intro}

Ceramic materials play a central role in modern joint replacement technology \cite{Biocompatibility}. One of the most widely used ceramics is BIOLOX\textsuperscript{\textregistered{}}\textit{delta}, an alumina matrix composite engineered by CeramTec GmbH for improved wear resistance and reduced fracture risk~\cite{Wearstudy,Yoon2018}. The material is valued for its excellent biocompatibility and extremely low wear rate (less than $0.1\,\text{mm}^3$ per million cycles)~\cite{Wearstudy,Biocompatibility}. Primary applications include femoral heads and acetabular liners, with increasing relevance in knee replacements~\cite{CeramicKnee,Biocompatibility}.
BIOLOX\textsuperscript{\textregistered{}}\textit{delta} implants undergo a complex, multi-step manufacturing process including powder preparation, forming, sintering, and precision machining. At each stage, specific defect types can arise—for example, cracks in the green body during pressing, microcracks from machining, internal material inhomogeneities or foreign materials. Such imperfections are critical, as they may significantly affect mechanical performance and often remain undetected until component failure.

To ensure quality, components are subjected to extensive mechanical testing. In particular, burst tests simulate extreme loading conditions by deliberately fracturing components to assess their failure behaviour, while proof tests apply high, non-destructive loads to verify structural integrity prior to clinical use. All fractures analysed in this study originate from such in-production burst and proof tests; no retrievals from clinical use were included. If failure occurs, fracture surfaces must be analyzed to determine the origin and cause of the defect. This process requires expert interpretation of microstructural features across various magnifications and is both time-consuming and potentially subjective. Fracture types are typically categorized based on their origin in the production chain, such as green body defects, hard machining defects, or material flaws. From a fractography perspective, experts first identify the \emph{fracture mirror}, a relatively smooth zone (transcrystalline fracture area) surrounding the initiation site—even in polycrystalline ceramics, where microstructural roughness is intrinsic~\cite{Quinn}. This region transitions into a rougher “mist” (intergranular fracture area) and then into hackle lines that radiate outward (so-called velocity hackles), guiding the eye back to the fracture origin~\cite{Quinn}. Accurate interpretation of morphological transitions is essential for identifying the fracture origin and differentiating competing failure mechanisms, providing a reliable root-cause analysis that safeguards implant quality and helps reduce revision surgeries.

In recent years, deep learning has become a powerful tool for microstructure analysis and defect inspection in materials science. Vision Transformers (ViTs), in particular, have shown strong results on small or noisy datasets due to their ability to model long-range dependencies without handcrafted features~\cite{whitman2025vit}, with successful applications to SEM-based defect classification in semiconductor manufacturing~\cite{IEEE2025_SemiconductorSEM} and to optical inspection of metallic surfaces~\cite{Alaa2024}.

Within fractography specifically, several CNN-based studies have addressed the analysis of fracture surfaces. Tsopanidis \textit{et al.}~\cite{TSOPANIDIS2020106992} introduced semantic segmentation of brittle fracture surfaces in oxide ceramics. Bastidas-Rodriguez \textit{et al.}~\cite{BASTIDASRODRIGUEZ2020104532} demonstrated fracture-mode classification in metallic materials, while more recent work has explored richer architectures and modalities, including transformer-based segmentation for morphological fractography~\cite{TANG2024110149}, and crack-feature segmentation that explicitly contrasts SEM imagery with topography data~\cite{SCHMIES2024107814}. Across these studies, however, the focus has been on \emph{descriptive} tasks (segmentation of fracture features, classification of fracture \emph{mode}) rather than on inferring the underlying \emph{root cause} along the manufacturing chain, and interpretability of model decisions is rarely addressed explicitly. Moreover, none of these works targets ceramic implant materials such as alumina matrix composites, where fractographic interpretation is not only a research question but also a regulated quality-assurance task.

The present study addresses this gap by designing, implementing, and evaluating an interpretable ViT workflow for automated classification of fracture \emph{causes} in an alumina matrix composite, combining multi-scale SEM evidence with Grad-CAM-based attribution that aligns with established expert reasoning.

\section{Materials and Methods}

\subsection{Dataset}\label{sub:dataset}
Fracture-analysis reports, created by multiple (human) ceramics experts, from 2019--2024 were supplied by CeramTec GmbH. Inter-operator agreement could not be retrospectively computed, but the labeling protocol was standardized. Each report documents one or more fractures, identifies the component by a fracture-origin analysis number (FOAN), and contains tables of metadata (component ID, fracture load, fracture type and sub-type) together with scanning electron microscopy (SEM) micrographs across magnifications ranging from 50× to 10\,000×.

Each FOAN typically corresponds to a \emph{multi-scale image set} of at least eight SEM images, capturing the visible part of the fracture origin on each relevant fracture fragment at progressively higher magnifications. Lower magnification views (50×–250×) provide an overview of the crack morphology, while higher magnifications (1\,000×–10\,000×) reveal fine-scale microstructural features such as pores, grain pull-out, and phase-boundary cracks.

To prevent spurious correlations, all SEM frames were cropped to remove scale bars and instrument text overlays. A random audit of 100 processed images confirmed that no magnification labels or frame annotations remained visible.

\paragraph{Image-report matching}
To build a supervised dataset, each image was matched to its report and fracture instance. OCR was applied to the PDF reports using Tesseract~5.3.4 to extract FOANs, serial numbers, and labels~\cite{OCREngine}. Image filenames were parsed via regular expressions to recover embedded FOANs and instance tags. Because FOANs were sometimes inconsistent and serial numbers were missing or ambiguous, a fuzzy-matching algorithm based on Levenshtein distance was implemented~\cite{coates2023identifyingdocumentsimilarityusing}. For each filename candidate, an exact FOAN match in the OCR table was sought first; if none was found, the closest FOAN with a similarity score above 0.9 was selected. If multiple matches remained, serial-number substrings were used to break ties. Matched images were assigned a unique identifier combining FOAN, serial number, and defect type; unmatchable files (\(\approx\)23\,\% of the archive) were discarded to preserve the integrity of the ground-truth annotations.

\begin{figure}[t]
    \centering
    \begin{tikzpicture}
    \begin{axis}[
        ybar,
        bar width=15pt,
        enlarge x limits=0.15,
        ymin=0,
        ylabel={Number of Images},
        ylabel style={yshift=-2mm},
        symbolic x coords={
            hard machining defect,
            green body defect,
            material defect,
            unknown origin,
            unmatchable
        },
        xtick={hard machining defect,green body defect,material defect,unknown origin,unmatchable},
        xticklabel style={rotate=45, anchor=east},
        nodes near coords,
        nodes near coords align={vertical},
        point meta=y,
        every node near coord/.append style={
            font=\footnotesize\bfseries,
            yshift=13pt,
            color=black,
            anchor=north
        },
        width=0.75\linewidth,
        height=6cm,
        title={Distribution of \texttt{fractureType} Labels},
        legend style={at={(0.99,0.99)}, anchor=north east, draw=none, fill=none},
        legend cell align={left}
    ]
    \addplot+[fill=blue!60] coordinates {
        (hard machining defect, 4351)
        (green body defect, 3787)
        (material defect, 355)
    };
    \addplot+[fill=orange!60] coordinates {
        (unknown origin, 275)
        (unmatchable, 2527)
    };
    \legend{Included in training, Excluded from training}
    
    \end{axis}
    \end{tikzpicture}
    \caption[Class distribution (cleaned dataset)]{Class distribution of the cleaned fracture dataset after image--FOAN matching.}
    \label{fig:fracturetype_distribution}
    \end{figure}
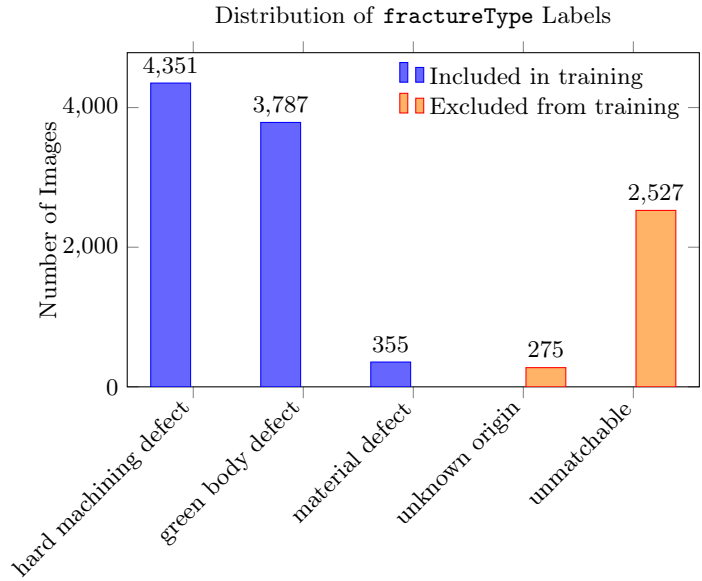

\paragraph{Dataset statistics}
The resulting dataset contained 8\,493 labelled SEM images.  Labels were aggregated into three classes: \emph{green body defect}, \emph{hard machining defect} and \emph{material defect}.  A fourth class, \emph{unknown origin}, and a set of unmatchable images were excluded from training because they lack a clear supervisory signal, however could be used in future work for semi-supervised or self-supervised approaches. Figure~\ref{fig:fracturetype_distribution} summarises the class distribution.  The dataset is highly imbalanced with the minority class assigned to only 4\% of images.  Magnifications were extracted via OCR of scale bars embedded in the original SEM frames.  Eight distinct magnifications were present (50$\times$, 100$\times$, 250$\times$, 500$\times$, 1k$\times$, 2k$\times$, 4k$\times$, 10k$\times$), with high magnification (\SI{1000}{\times}--\SI{10 000}{\times}) used more frequently in expert analyses than low magnification (\SI{50}{\times} -- \SI{100}{\times}).

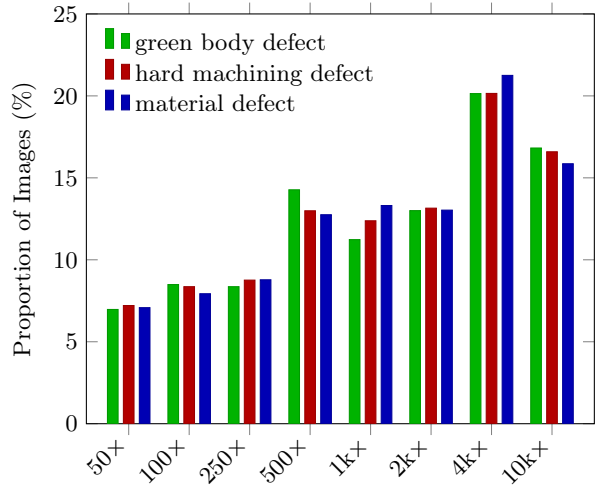
\begin{figure}[h]
\centering
\begin{tikzpicture}
\begin{axis}[
    ybar,
    width=0.75\linewidth,
    height=7cm,
    ymin=0,
    ymax=25,
    ylabel={Proportion of Images (\%)},
    symbolic x coords={50×,100×,250×,500×,1k×,2k×,4k×,10k×},
    xtick=data,
    xticklabel style={rotate=45, anchor=east},
    bar width=4pt,
    x=0.8cm,
    legend style={
        at={(0.02,0.98)},        
        anchor=north west,       
        legend columns=1,        
        font=\small,             
        draw=none,               
        fill=none                
    },
    legend cell align={left}
]

\addplot+[ybar, fill=green!70!black, draw=green!60!black, bar shift=-6pt] 
    coordinates {(50×,6.97) (100×,8.49) (250×,8.36) (500×,14.27) (1k×,11.23) (2k×,13.00) (4k×,20.14) (10k×,16.82)};
\addlegendentry{green body defect}

\addplot+[ybar, fill=red!70!black, draw=red!60!black, bar shift=0pt] 
    coordinates {(50×,7.21) (100×,8.36) (250×,8.76) (500×,12.99) (1k×,12.38) (2k×,13.15) (4k×,20.15) (10k×,16.59)};
\addlegendentry{hard machining defect}

\addplot+[ybar, fill=blue!70!black, draw=blue!60!black, bar shift=6pt] 
    coordinates {(50×,7.08) (100×,7.93) (250×,8.78) (500×,12.75) (1k×,13.31) (2k×,13.03) (4k×,21.25) (10k×,15.86)};
\addlegendentry{material defect}

\end{axis}
\end{tikzpicture}
\caption{Magnification distribution per fracture type (proportions within class).}
\label{fig:mag_class_bar}
\end{figure}

In addition to class counts, the distribution of SEM magnifications within each fracture type was analysed (Fig.~\ref{fig:mag_class_bar}).
The proportions are broadly similar across classes. The balanced distribution reduces the risk that the classifier could exploit magnification as a spurious shortcut, ensuring that morphological features rather than scale dominate the decision process.

\subsection{Model architecture and training}
A ViT‑Base‑Patch16‑224 architecture was fine-tuned using PyTorch for model implementation and training, and Hydra for configuration management \cite{Paszke,Hydra}.  The model was initialized with weights pre‑trained on ImageNet‑21k and fine‑tuned on ImageNet‑1k \cite{Wightman}.  Further fine‑tuning on the micrographs dataset employed the AdamW optimiser with cosine‑ annealing learning‑ rate scheduling.  Hyperparameters were optimised using a Tree- Structured Parzen Estimator via Optuna~\cite{akiba2019optuna} across learning rate, weight decay, label smoothing, stochastic depth and data‑ augmentation parameters with the goal to maximize the macro-averaged $F_1$ score. The explored hyperparameter search space is summarised in the supplementary material (available at \url{https://github.com/Nailujj/FractoViT}). Search ranges were selected to balance computational feasibility with coverage of key architectural and optimisation parameters known to affect Vision Transformer fine-tuning, including learning-rate scaling across layers, stochastic depth, and data-augmentation intensity.  Class imbalance was addressed by (i) constructing a weighted random sampler proportional to the inverse class frequencies, ensuring balanced batches during training, and (ii) employing focal loss with focusing parameter \(\gamma=2\) to down‑weight easy examples and emphasise hard, minority‑class samples.  Extensive data augmentation was applied using RandAugment (tunable operations per image with magnitude sampled in \([5,15]\)) to enhance generalisation.

\subsection{Training and evaluation}\label{subsec:evaluation}

Due to the limited availability of minority-class samples, a small, fully independent hold-out test set was only employed for qualitative assessment. The model's quantitative generalisation and performance were rigorously assessed using a stratified five-fold cross-validation (CV) scheme. Hyperparameter optimisation was performed based on validation set performance. Performance was evaluated by averaging macro-$F_1$, accuracy, precision, and recall across the five folds. Additionally, Gradient-weighted Class Activation Mapping (Grad-CAM)~\cite{Selvaraju2017_GradCAM} heatmaps were generated for a hold-out batch of fractured components (collected after the 2019-2024 archive period and unseen during training or validation) to qualitatively assess model predictions and identify key morphological features driving classification decisions. 

\paragraph{Cross-validation splitting and leakage audit}
Stratified five-fold cross-validation was performed at the \emph{image level} rather than at the specimen (FOAN) level. This choice maximises the number of training samples per fold, which is critical given the severe imbalance of the material-defect class (only 355 images). However, since each FOAN typically contributes multiple SEM micrographs taken at different magnifications of overlapping fracture regions, an image-level split could in principle allow morphologically similar views of the same specimen to occur in both training and validation folds and thereby inflate the reported metrics—a known risk in medical and materials imaging where subject-wise splits are generally preferred~\cite{HowYouSplitMatters}. To quantify and bound this risk \emph{post hoc}, a two-stage similarity audit was performed.

First, perceptual hashing (pHash) was used to retrieve, for each validation image, the top-$K$ most similar training images within a Hamming distance $\leq 64$ ($K = 50$). 
This step ensures high recall of potential duplicates, even across magnifications, because pHash encodes coarse perceptual structure rather than exact pixel values. 
Second, candidate pairs were verified using the Structural Similarity Index (SSIM; grayscale, maximum side 128\,px, \texttt{data\_range=255}), which measures luminance, contrast, and structural correlation and is widely used in image quality assessment~\cite{Wang}. 
The proportion of validation images with at least one SSIM-confirmed near-duplicate in the training set was computed for thresholds from 0.50 to 0.95.

As shown in Table~\ref{tab:ndcheck}, leakage rates dropped sharply with stricter SSIM criteria and remained consistently below 1\,\% for SSIM~$\geq$~0.80 across all folds, indicating that near-duplicate overlap between training and validation folds is negligible and that the reported cross-validation performance is therefore only marginally affected by within-specimen leakage.

\begin{table}[h]
\centering
\caption{Proportion of validation images (\%) with at least one near-duplicate in the training set, confirmed by SSIM at different thresholds.}
\label{tab:ndcheck}
\begin{tabular}{lccccccc}
\hline
Fold & 0.50 & 0.60 & 0.70 & 0.80 & 0.85 & 0.90 & 0.95 \\
\hline
1 & 16.64 & 6.83 & 1.97 & 0.79 & 0.79 & 0.79 & 0.62 \\
2 & 14.56 & 6.10 & 1.81 & 0.45 & 0.34 & 0.34 & 0.28 \\
3 & 14.84 & 6.26 & 1.92 & 0.56 & 0.45 & 0.45 & 0.45 \\
4 & 16.42 & 6.04 & 2.26 & 0.90 & 0.73 & 0.51 & 0.45 \\
5 & 15.75 & 5.64 & 2.09 & 0.68 & 0.56 & 0.45 & 0.40 \\
\hline
\end{tabular}
\end{table}

\paragraph{Qualitative expert assessment}
The quality of the resulting attributions was rated in a structured consensus session by three in-house fractography experts with several years of routine experience in ZTA failure analysis. For each hold-out fracture instance, the experts were jointly presented with the original SEM micrograph, the predicted class with its softmax confidence, the ground-truth label, and the corresponding Grad-CAM heat-map across all available magnifications. They then discussed (i) whether the highlighted regions coincided with morphological cues that a human analyst would consider diagnostic for the predicted class (e.g.\ fracture mirror, hackle field, pore, machining mark), (ii) whether the model's reasoning appeared mechanistically plausible, and (iii) whether any attribution constituted a \emph{right-for-the-wrong-reason} pattern. Disagreements were resolved through discussion until a unanimous joint rating was reached for each instance. No formal inter-rater agreement statistic was computed because ratings were generated by consensus rather than independently; this is acknowledged as a limitation.

\section{Results \& Discussion}\label{sec:results_discussion}

\subsection{Hyperparameter optimisation and model selection}
The hyperparameter search was conducted over 100 trials of 20~epochs each, with under-performing configurations pruned early. 
The explored search space and best performing configuration is summarised in the supplementary material (available at \url{https://github.com/Nailujj/FractoViT}). 
The best-performing configuration employed a base learning rate of \(7.3\times10^{-5}\), weight decay of 0.14, layer-wise learning-rate decay of 0.843, and two RandAugment operations of magnitude~12, without label smoothing. 
This combination achieved a validation macro-F\(_1\) of \textbf{0.86} prior to cross-validation. 
Increasing regularisation through stronger stochastic depth, patch drop, or more aggressive augmentation did not yield further improvements, indicating that the observed generalisation gap (Section~\ref{subsec:evaluation}) is unlikely to result from insufficient regularisation. 
Instead, it is more plausibly attributed to residual label noise and inconsistencies in the historical dataset, including fuzzy-matched entries and variability in human annotations. This interpretation is consistent with the quantitative results presented in the following section.

\subsection{Cross-validation performance}

\begin{table}[h]
    \centering
    \caption{
        Five-fold cross-validation results for the ViT model.  
        Accuracy, precision, recall, and macro-F$_1$ are computed on the validation splits.  
        Train macro-F$_1$ refers to the same epoch with the highest validation macro-F$_1$ in each fold.  
        $\Delta F_1$ denotes the generalisation gap (train–val).
    }
    \label{tab:cv}
    \begin{tabular}{lccc}
        \hline
        Metric & Mean $\pm$ SD & 95\,\% CI \\
        \hline
        Accuracy (val)     & 0.907 $\pm$ 0.008 & [0.897, 0.917] \\
        Macro-F$_1$ (val)  & 0.888 $\pm$ 0.017 & [0.867, 0.909] \\
        Macro-F$_1$ (train)& 0.998 $\pm$ 0.001 & [0.997, 0.999] \\
        $\Delta F_1$ (train–val) & 0.090 $\pm$ 0.012 & [0.075, 0.105] \\
        \hline
    \end{tabular}
\end{table}

Table~\ref{tab:cv} reports the mean validation metrics over the five cross-validation folds.
For completeness, the corresponding training metrics for the same epochs and the aggregated learning curves are provided in the Supplementary Material (available at \url{https://github.com/Nailujj/FractoViT}).
On average, the model achieves a macro-F$_1$ of \textbf{0.888 $\pm$ 0.017} and accuracy of \textbf{0.907 $\pm$ 0.008 } on the validation sets.
The generalisation gap between training and validation macro-F$_1$ is approximately \textbf{9 percentage points}, indicating moderate overfitting.
Given the pronounced class imbalance and the inclusion of noisy or weakly-labelled samples from the fuzzy-matching stage, such a gap is expected.
Moreover, the stability of validation performance across folds and the negligible leakage rates (Section~\ref{subsec:evaluation}) suggest that the gap does not stem from memorisation of specimen-specific features, but from intrinsic label noise and minority-class scarcity.

The qualitative findings are broadly comparable to those reported for ViT-based defect classification in other industrial domains.
Alaa et al. \cite{Alaa2024} achieved 93.5 \% accuracy on metallic surface defects with optical microscopy images, while Huang et al. \cite{IEEE2025_SemiconductorSEM} reported over 90 \% accuracy for semiconductor wafer defect detection on SEM images.
However, both studies used datasets without the hierarchical, multi-scale acquisition structure inherent to fracture-origin analysis. Contrary, in the present work, the dataset exhibits three compounding challenges:
(i) multi-scale image sets per fracture, spanning magnifications from 50× to 10 000×,
(ii) severe class imbalance (10:1 between majority and minority classes), and
(iii) label noise introduced by fuzzy matching of historical records.
Despite these constraints, the ViT achieves 90.7 \% accuracy, underscoring the model’s robustness and the transferability of attention-based architectures to multi-scale fracture-surface analysis.

\paragraph{Error structure}

\begin{figure} [h]
    \centering
    \includegraphics[width=0.5\linewidth]{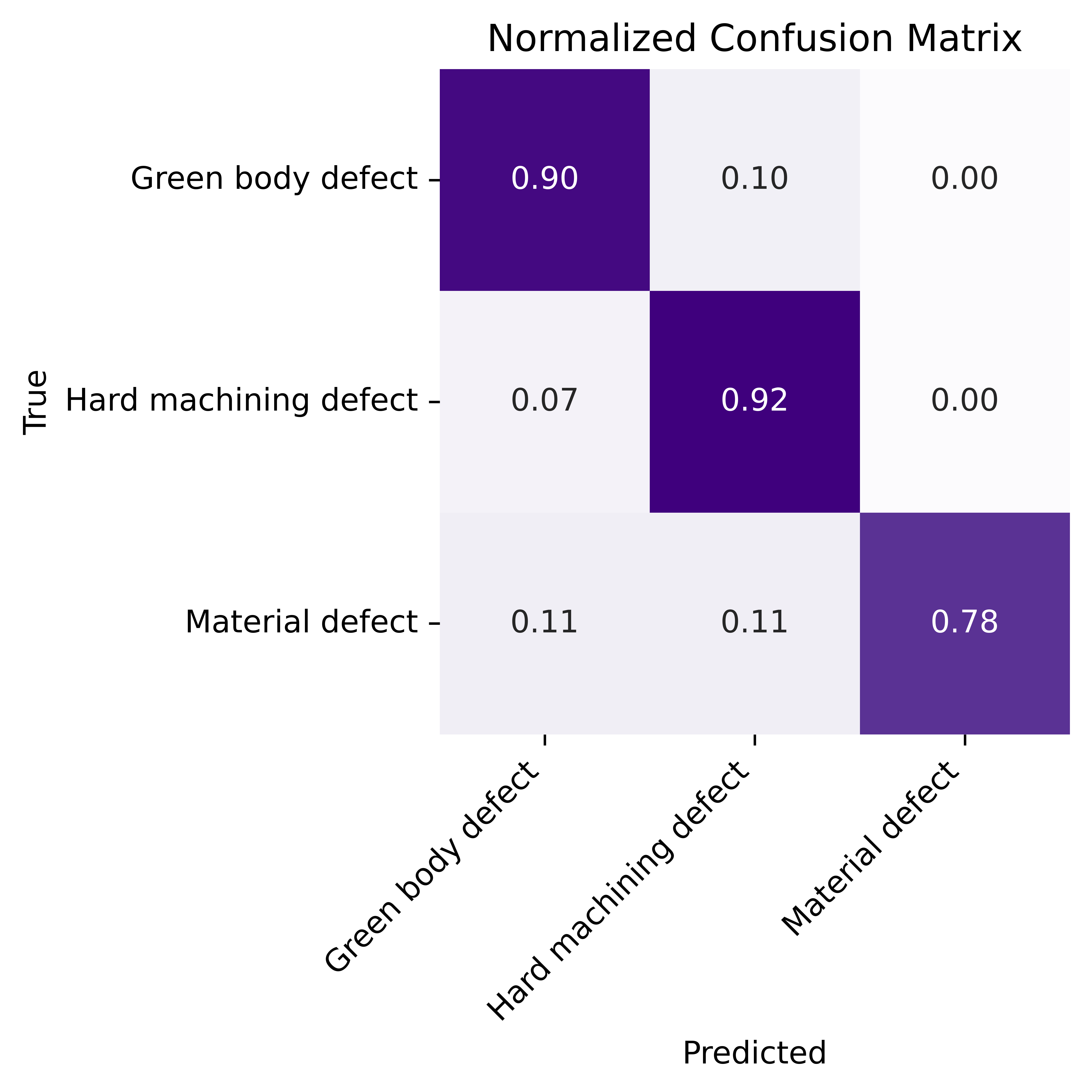}
    \caption{Aggregated, row-normalised confusion matrix over the five cross-validation folds.}
    \label{fig:confusion}
\end{figure}

The confusion matrix in Fig.~\ref{fig:confusion} reveals a largely
\emph{symmetrical} confusion between \emph{green body} and \emph{hard machining}
defects: 10\,\% of green body images are mis-labelled as hard machining, while
7\,\% of hard machining samples are assigned the opposing class.  
For the \emph{material-defect} class the pattern is asymmetric: 22 \% of
material-defect images leak into the other two classes (11 \% each), limiting
recall to~0.78, \emph{yet} almost no samples from the other classes are
mistakenly predicted as material defect, yielding a column-wise precision of
0.94.   This behaviour is consistent with the visual similarity of early-process cracks
(green body vs.\ machining) and the relative scarcity of material-defect
examples, which makes the classifier conservative—i.e.\ high precision but lower
recall—when assigning that label.

\paragraph{Per-class metrics}
Table~\ref{tab:perclass} quantifies the pattern: recall for \emph{material defects} is lowest (0.777) owing to limited training samples, but its precision peaks at 0.939 because the classifier only assigns the label when attention maps exhibit clear pore or phase-boundary cues.  
The interplay of focal loss and a weighted sampler thus trades recall for precision in the rare class—a desirable bias when false positives are costlier than false negatives in root-cause analysis.
\begin{table}[h]
    \centering
    \caption{Per‑class performance derived from the aggregated confusion matrix over five folds.}
    \label{tab:perclass}
    \begin{tabular}{lccc}
        \hline
        Class & Recall & Precision & F1 \\
        \hline
        Green‑body defect & 0.898 & 0.903 & 0.901 \\
        Hard‑machining defect & 0.924 & 0.907 & 0.916 \\
        Material defect & 0.777 & 0.939 & 0.851 \\
        \hline
    \end{tabular}
\end{table}

\subsection{Effect of magnification}

\begin{figure}[h]
    \centering
    \includegraphics[width=0.75\textwidth]{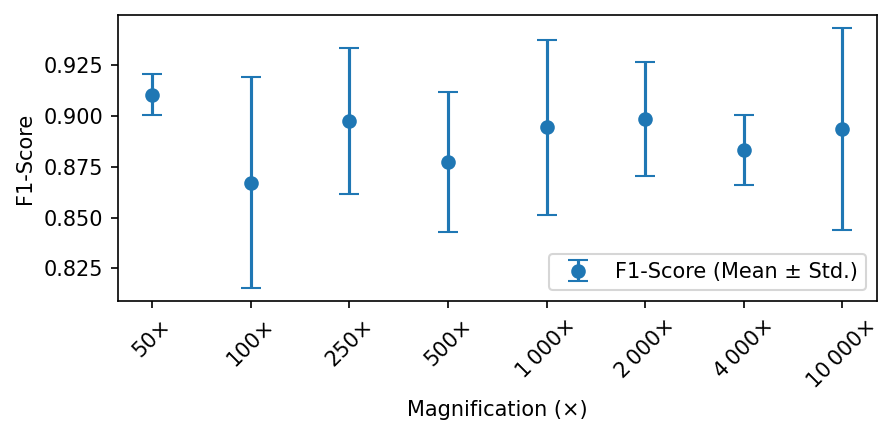} 
    \caption{$F_1$ score of the ViT model across magnifications. Error bars denote standard deviation across folds. Robust performance at low magnification indicates that macro-scale fracture patterns are predictive of underlying causes. \emph{Mechanistic note:} macro-scale mirror geometry and hackle-field organisation appear sufficient to separate manufacturing-related fracture origins in many cases.}
    \label{fig:magperf}
\end{figure}

Although high magnification is typically considered essential for fracture analysis (\cite{Quinn2020Fractography}), images at 50× achieve F$_1$ scores within one standard deviation of those at 4k–10k×, suggesting comparable diagnostic value (Fig.~\ref{fig:magperf}); a formal statistical comparison was beyond the scope of this work. This indicates that macro-scale fracture patterns—radial fans, arrest lines—already encode enough statistical regularities for the ViT to diagnose the origin, challenging the conventional analyst workflow that prioritises high-magnification inspection first. In practice, an automated pre-screening step at low magnification could therefore accelerate routine failure analysis by flagging the most probable cause before expensive high-mag imaging. Mechanistically, this observation can be rationalised by classical fractography: global features such as mirror extent, shape, and hackle-field organisation are known to reflect local crack velocity gradients and stress fields around the initiation site~\cite{Quinn}. These cues differ systematically between manufacturing-related defect types, explaining why low-magnification views already provide sufficient diagnostic signal.

\subsection{Qualitative model interpretation}
The examples discussed in this section were selected from the consensus-rated hold-out batch (Section~\ref{subsec:evaluation}) as representative of the patterns identified by the expert panel.

\begin{figure*}[t] 
  \centering
  \begin{subfigure}[b]{0.3\textwidth}
    \centering
    \includegraphics[width=\textwidth]{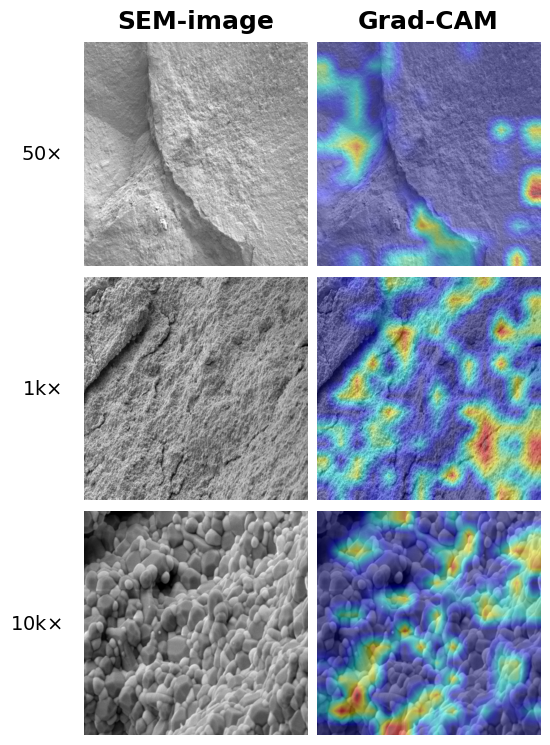}
    \caption{Green body defect}
    \label{fig:gbd}
  \end{subfigure}
  \hfill
  \begin{subfigure}[b]{0.3\textwidth}
    \centering
    \includegraphics[width=\textwidth]{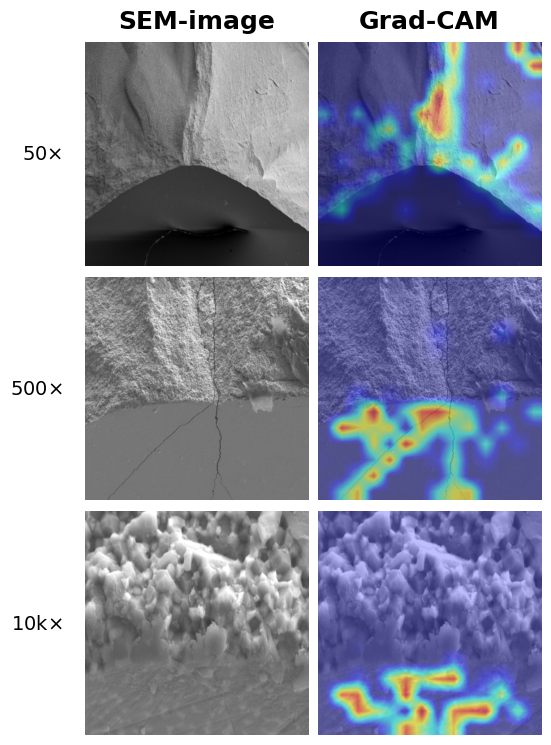}
    \caption{Hard machining defect}
    \label{fig:hmd}
  \end{subfigure}%
  \hfill
  \begin{subfigure}[b]{0.3\textwidth}
    \centering
    \includegraphics[width=\textwidth]{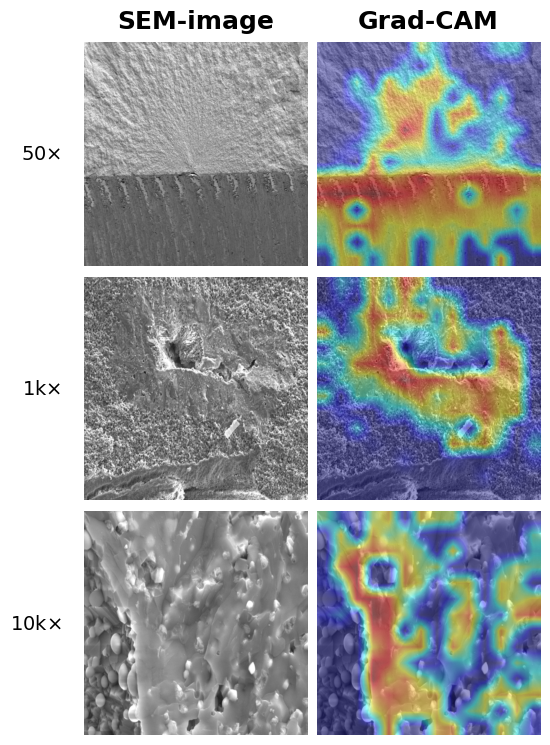}
    \caption{Material defect}
    \label{fig:md}
  \end{subfigure}

  \caption{Grad-CAM visualisation of the three defect classes (one fracture instance each) at increasing magnifications. For each class the original SEM image is shown on the left and the corresponding Grad-CAM heat-map on the right.  All panels were correctly classified by the model with confidences above 90 \%.}
  \label{fig:three-subfigures}
\end{figure*}

Figure~\ref{fig:three-subfigures} illustrates representative Grad-CAM visualisations from the hold-out batch for each defect class across increasing SEM magnifications. In all three cases the highlighted regions coincide with morphological cues that are physically meaningful in fracture analysis, such as fracture mirrors, hackle lines, grain boundary structures, and machining traces.

\paragraph{Green body defects}
At low magnification (Fig.~\ref{fig:gbd}, 50$\times$) the model emphasizes edges in the fracture mirror and shading variations around the origin. These structures are consistent with the geometry of hackle lines, which typically point back to the initiation site. While a definitive diagnosis is rarely possible at this scale, the model already extracts useful information and predicts the defect correctly. At intermediate magnification (1k$\times$), intergranular cracks and shadow-like grain boundary features become prominent, and the model highlights exactly these regions. At the highest magnification (10k$\times$) it focuses on rounded grain surfaces that are characteristic of pre-sintering defects. This progression across scales mirrors the logic of established fractographic reasoning, making the model's decision highly plausible.

\paragraph{Hard machining defects}
In the case of machining-induced defects (Fig.~\ref{fig:hmd}), the model relies at 50$\times$ on fine surface lines that are not visually striking, yet its prediction is correct. At higher magnifications, breakout regions and cracks are emphasised, which are robust indicators of machining damage. At the highest scale, machining marks converging toward the origin are strongly attended, providing circumstantial evidence of the failure mechanism. Across magnifications, the model's focus consistently aligns with meaningful morphological cues for machining defects.

\paragraph{Material defects}
For material defects (Fig.~\ref{fig:md}), the low- magnification view at 50$\times$ provides only weak diagnostic cues. The model predominantly attends to machining marks, metallic abrasion, and the fracture-mirror, while the actual defect remains largely undetected. Nevertheless, it predicts a material defect— likely because the metallic residues are interpreted as foreign material. While this interpretation is understandable, these residues are not the fracture origin. This constitutes a \emph{right-for-the-wrong-reason} pattern that, despite the nominally correct label, limits the trustworthiness of low-magnification predictions for the material-defect class and motivates expert verification at higher magnifications. At higher magnifications (1k$\times$ and 10k$\times$), pore-related irregularities and adjacent grain connections become visible, and the model shifts its attention accordingly. In these cases the decision is well supported by features that are genuinely diagnostic of intrinsic material flaws.

\begin{figure}
    \centering
    \includegraphics[width=0.3\linewidth]{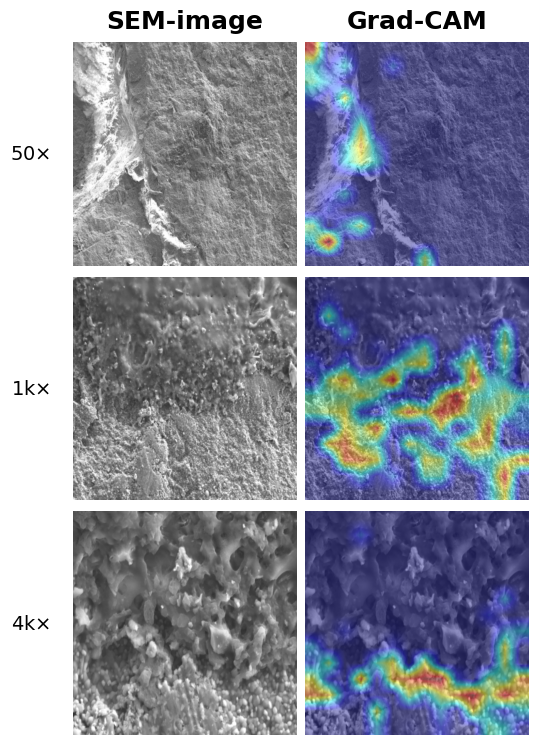}
      \caption{Example of a misclassification. In each case, a \emph{green body defect} (ground-truth label) is erroneously predicted as a \emph{hard machining defect} within the same fracture instance. For each magnification-level, the left column shows the raw SEM image and the right column the corresponding Grad-CAM heat-map.}
    \label{fig:HardCAM}
\end{figure}

\paragraph{Misclassifications}
Figure~\ref{fig:HardCAM} illustrates typical error cases where the model predicts a machining defect instead of the correct green-body defect. The underlying problem is that the images contain fracture mirrors but not the actual initiation site. At 50$\times$, the model highlights structures resembling machining marks, which is misleading in the absence of hackle lines or origin features. At higher magnifications, carbon residues obscure the surface, further reducing interpretability. These examples underline that the model can produce plausible but ultimately incorrect attributions when the true fracture origin is missing or imaging conditions are compromised.

\paragraph{Summary}
Taken together, the Grad-CAM analyses show that the model predominantly relies on physically meaningful cues: fracture mirrors, hackle lines, pores, and machining marks. At low magnification it already extracts information that is not always obvious by eye, whereas at higher magnifications the attended regions match established diagnostic criteria very well. Limitations arise primarily when the fracture origin is not present in the field of view or when SEM artefacts such as carbon residues obscure relevant features. In such cases, the model's predictions can appear plausible but are not trustworthy. Overall, the results indicate that the model can serve as a complementary tool: accelerating analysis through pre-screening at low magnification, while definitive classification remains with expert confirmation at higher scales.

\subsection{Mechanistic insights from model attention and magnification effects}
Across magnifications, model attention converged on canonical fractographic features—fracture mirrors, mist-to-hackle transitions, pores, and machining marks, linking the predictions to mechanisms that are routinely used by human experts. Notably, the near-equality of performance at low (50×) and high (1k×–10k×) magnification indicates that \emph{macro-scale fracture morphology} already encodes sufficient diagnostic signal. From a mechanistic perspective, the geometry of the mirror and the spatial organisation of hackle line fields appear to capture propagation velocity gradients and local stress concentrations around the initiation site, which differ systematically between green-body, hard machining, and intrinsic material origins. This interpretation is consistent with established fractography and explains why reliable classification is possible without resolving fine-scale microstructural details in many cases.

The observed precision–recall asymmetry for the \emph{material-defect} class (high precision, lower recall) further suggests that pore-associated or phase-boundary cues are highly specific but sparsely expressed in the available imagery. In practical terms, the model behaves conservatively when assigning ``material defect''—a desirable bias when false positives would trigger costly root-cause actions in quality control.

\paragraph{Implications for quality control}
By exploiting macro-scale morphology, an interpretable low-magnification pre-screen can prioritise samples for high-magnification confirmation, reducing microscope time and standardising triage decisions. Because Grad-CAM highlights overlap with classical cues, the workflow provides traceable \emph{mechanistic rationales}, facilitating expert acceptance and auditability in regulated environments.

\section{Conclusion}
An interpretable vision-transformer workflow was introduced for multi-scale classification of fracture causes in ZTA ceramics. Beyond aggregate performance (accuracy 0.907; macro-F$_1$ 0.888 under severe class imbalance), the study provides \emph{mechanistic insight}: macro-scale fracture morphology—in particular mirror geometry and the organisation of hackle line fields—contains surprisingly rich diagnostic information that distinguishes manufacturing-related fracture origins. Grad-CAM analyses demonstrate that model decisions co-localise with canonical fractographic features (mirrors, hackles, pores, machining marks), yielding traceable rationales that align with expert reasoning and can be audited. Future work should explore (i) \emph{multiple-instance learning} across full multi-image fracture sets, (ii) \emph{cross-modal fusion} with optical microscopy to reduce high-magnification dependence, and (iii) \emph{pre-training on unmatched or weakly labeled images} to mitigate class imbalance and leverage the large volume of uncurated data.

\paragraph{Limitations and generalisability}
A potential limitation is the employment of fuzzy-matching techniques to a small subset of the images in the image-report matching process, which, even though affected <10\% of images, could be one of the reasons for the train/val gap described in \ref{sec:results_discussion}.
While the hold-out and validation sets support robustness of the interpretability findings, the data originate from a single production environment; broader external validation across materials, sites and instruments remains an important next step. Further, a FOAN-grouped split would constitute the methodological gold standard and is recommended for future work, particularly once the minority-class sample size permits stratification at the specimen level without compromising training-set coverage.

\FloatBarrier

\section*{Declaration of competing interest}
All authors are employees of CeramTec GmbH, the manufacturer of BIOLOX\textsuperscript{\textregistered{}}\textit{delta}. This relationship is disclosed for transparency. The authors declare that they have no other known competing financial interests or personal relationships that could have appeared to influence the work reported in this paper.

\section*{Data availability}
Due to contractual restrictions, the raw scanning electron microscopy (SEM) images cannot be made publicly available at this point. While the material's general composition is well-documented, the collective analysis of these images could provide a proprietary understanding of the microstructure and material properties that are central to the implant's performance. The complete source code is available at \url{https://github.com/Nailujj/FractoViT}.

\section*{Acknowledgements}
The authors thank Daniel Behr, Lecturer at the University of Applied Sciences and Arts Northwestern Switzerland, for supervising the Bachelor's thesis that laid the foundation for this study and for his valuable feedback on the methodology and results.

The initial model was developed by Julian Schmid during his Bachelor's thesis, which was carried out at CeramTec GmbH as part of an industry collaboration with the University of Applied Sciences and Arts Northwestern Switzerland. All infrastructure, data, and domain expertise were provided by CeramTec GmbH.

The authors also acknowledge the use of large language models to support literature exploration and the drafting of a small number of text passages, which were subsequently reviewed and adapted by the authors. All analyses, interpretations, and conclusions are the authors’ own.

\section*{CRediT authorship contribution statement}
\textbf{Julian Schmid:} Conceptualization, Methodology, Quantitative validation, Visualization, Software, Formal analysis, Writing – original draft.
\textbf{Pawel Astankow:} Data curation, Writing – review and editing.
\textbf{Tom Vater:} Qualitative validation, Writing – original draft.
\textbf{Julius Beck:} Quantitative validation, Writing – review and editing.
\textbf{Robert Cichon:} Supervision, Writing – review and editing.
\textbf{Danny Krautz:} Supervision, Project administration.


\bibliographystyle{splncs04}
\bibliography{paper}     
\FloatBarrier

\end{document}